\documentclass[conference]{IEEEtran}
\IEEEoverridecommandlockouts
\usepackage{cite}
\usepackage{amsmath,amssymb,amsfonts}
\usepackage{algorithmic}
\usepackage{graphicx}
\usepackage{textcomp}
\usepackage{xcolor}
\usepackage{hyperref}
\usepackage{balance}
\hypersetup{
    colorlinks=true,
    linkcolor=blue,
    filecolor=magenta,      
    urlcolor=blue,
}
\def\BibTeX{{\rm B\kern-.05em{\sc i\kern-.025em b}\kern-.08em
    T\kern-.1667em\lower.7ex\hbox{E}\kern-.125emX}}
    
\title{AI Learns to Recognize Bengali Handwritten Digits: \\Bengali.AI Computer Vision Challenge 2018\\
}

\author{
\IEEEauthorblockN{Sharif Amit Kamran}
\IEEEauthorblockA{
\textit{Bengali.AI}\\
Dhaka, Bangladesh \\
amit@bengali.ai}
\and
\IEEEauthorblockN{Ahmed Imtiaz Humayun}
\IEEEauthorblockA{
\textit{Bengali.AI}\\
Dhaka, Bangladesh \\
imtiaz@bengali.ai}
\and
\IEEEauthorblockN{Samiul Alam}
\IEEEauthorblockA{
\textit{Bengali.AI}\\
Dhaka, Bangladesh \\
samiul.a@bengali.ai}
\and
\IEEEauthorblockN{Rashed Mohammad Doha}
\IEEEauthorblockA{
\textit{Bengali.AI}\\
Dhaka, Bangladesh \\
rashed@bengali.ai}
\and
\IEEEauthorblockN{Manash Kumar Mandal}
\IEEEauthorblockA{
\textit{Bengali.AI}\\
Dhaka, Bangladesh \\
manash@bengali.ai}
\and
\IEEEauthorblockN{Tahsin Reasat}
\IEEEauthorblockA{
\textit{Bengali.AI}\\
Dhaka, Bangladesh \\
reasat@bengali.ai}
\and
\IEEEauthorblockN{Fuad Rahman}
\IEEEauthorblockA{
\textit{Apurba Technologies}\\
Sunnyvale, California \\
fuad@apurbatech.com}
}

\begin{document}
\maketitle

\maketitle

\begin{abstract}
Solving problems with Artificial intelligence in a competitive manner has long been absent in Bangladesh and Bengali-speaking community. On the other hand, there has not been a well structured database for Bengali Handwritten digits for mass public use. To bring out the best minds working in machine learning and use their expertise to create a model which can easily recognize Bengali Handwritten digits, we organized Bengali.AI Computer Vision Challenge. The challenge saw both local and international teams participating with unprecedented efforts.
\end{abstract}


\section{Introduction}
It has been a long time since any well structured Bengali handwritten digits dataset have been open sourced for research purpose. Though there is many handwritten dataset for Arabic numerals like MNIST \cite{mnist}, French \cite{french}, Chinese \cite{chinese}, Urdu \cite{urdu}, a massive void was present in terms of Bengali handwritten digits dataset, let alone characters. Several attempts were made to accumulate such dataset like the one collected by Indian Statistical Institute (ISI) \cite{isi}, but only a small amount was released for public access. Another such attempt was done by Center for Microprocessor Application for Training Education and Research (CMATER) group \cite{cmater}, but it was kept for internal research purposes only. In 2018, NumbtaDB \cite{numtadb} was released, which is the first of its kind, a well structured dataset for Bengali handwritten digits. Collected from multiple sources and open sourced for mass public use, the 85,000+ digits dataset is the largest such dataset for Bengali handwritten digits. Most surprising thing about the dataset was that all the cleaning, standardization and sorting was done by hand which occurred in a span of 5 months.

The natural next step was to organize a competition with this large scale data to connect and bring out the best of the machine learning practitioners of the country. The Bengali.AI CV challenge commenced in the month of June. A grand total of 57 teams participated from 26 local and international institutions ranging from college and universities. The overall number of submissions was around 650, peaking at 100 on the last week of the deadline. The challenge was hosted on Kaggle \cite{kaggle} which enabled the teams to use and exploit it's computing powers and subsystem for processing data and executing algorithms. Furthermore, the team could use the discussion forums for any queries or problems they had faced. The Kaggle kernels enabled both the veterans to teach the new machine learning practitioners and also made way for the participating teams to share their winning algorithms.

Most of the algorithms tried to address not only the recognition of digits but also the orientation and style of how the digits were written. The data was sorted in such a way that it contained different natural settings which can easily confuse a hard coded algorithm while keeping it recognizable by humans. The challenge was to produce such sophisticated machine learning algorithms which will be at par with their human counterparts. And the winning solutions were successful to produce such models by training on large amount of heavily augmented and variable datasets.

This competition is the first of a series of computer vision challenges hosted by Bengali.AI. The aim is to create a large artificial intelligence and machine learning community in Bangladesh similar to the  community created by computer vision challenges like Imagenet \cite{imagenet}, Pascal-VOC \cite{voc}, MS-COCO \cite{coco} etc.  which have become the cornerstone for participation and knowledge sharing in computer vision. Frequent challenges and competition will also pave the way for machine learning community in Bangladesh to grow and prosper and will motivate them to take part in international challenges. 
 
\section{Data Accumulation and Structuring}
\subsection{Data from Different Sources}

The data was gathered from six different sources as mentioned in \cite{numtadb}. The first source was the Bengali Handwritten Digits Database (BHDDB) which initially consisted of 23,400 samples. Later 209 samples were removed from it as they were illegible. It was contributed by Bangladesh University of Engineering and Technology's students from Department of Computer Science and Engineering. The data was collected in a form which had a distinct grid pattern. It was collected in RGB format.

The second data was also collected from BUET. The dataset is called BUET101 Database (B101DB) which had 435 samples of which 7 samples were removed. The data was cropped and labelled by hand after collecting.

OngkoDB was yet another contribution from Department of Computer Science and Engineering from BUET, which consisted of 28,900 samples. From this set, 321 number of samples were removed. The image format was Gray-scale, which was different from other collected data.

Students from Institute of Statistical Research and Training (ISRTHDB), Dhaka University, contributed the fourth data. It consisted of 13,133 samples of which 277 were rejected. It had the same format as the BHDDB dataset. Though it was much more pristine and less noisy. 

The fifth dataset was BanglaLekha-Isolated \cite{banglalekha} which consisted of  numerals and alphabets combined. Of which, we selected the numerals only for digit classification. The data was cleaned thoroughly and erroneous labels were fixed after accumulating. The total number of samples were 20,319 and was collected in Binary format. 572 out of the original samples were removed.

UIUDB, a dataset prepared by students from United International University was the last samples we collected for the competition. It had 576 samples of which 81 were removed. 

Table \ref{tab1} shows contribution by different sources and the number of samples which were included in the dataset.

\begin{table}[htbp]
\caption{Data Sources}
\begin{center}
\begin{tabular}{llllll}
\cline{1-5}
\multicolumn{1}{|l|}{\begin{tabular}[c]{@{}l@{}}Name of \\ the Dataset\end{tabular}}& \multicolumn{1}{l|}{\begin{tabular}[c]{@{}l@{}}Contributing \\ Institution\end{tabular}}& \multicolumn{1}{l|}{\begin{tabular}[c]{@{}l@{}}Number of \\ Samples\end{tabular}}& \multicolumn{1}{l|}{\begin{tabular}[c]{@{}l@{}}Removed \\ samples\end{tabular}}& \multicolumn{1}{l|}{\begin{tabular}[c]{@{}l@{}}File \\ format\end{tabular}}& \\ \cline{1-5}
\multicolumn{1}{|l|}{\begin{tabular}[c]{@{}l@{}}Bengali \\ Handwritten \\ Digits  Database \\ (BHDDB)\end{tabular}} & \multicolumn{1}{l|}{\begin{tabular}[c]{@{}l@{}}CSE, \\ BUET\end{tabular}}& \multicolumn{1}{l|}{23,400}& \multicolumn{1}{l|}{209} & \multicolumn{1}{l|}{RGB}&  \\ \cline{1-5}
\multicolumn{1}{|l|}{\begin{tabular}[c]{@{}l@{}}BUET101 \\ Database\\ (B101DB)\end{tabular}}                       & \multicolumn{1}{l|}{BUET}                                                                & \multicolumn{1}{l|}{435}                                                          & \multicolumn{1}{l|}{7}                                                          & \multicolumn{1}{l|}{RGB}                                                    &  \\ \cline{1-5}
\multicolumn{1}{|l|}{OngkoDB}                                                                                      & \multicolumn{1}{l|}{CSE, BUET}                                                           & \multicolumn{1}{l|}{28,900}                                                        & \multicolumn{1}{l|}{321}                                                        & \multicolumn{1}{l|}{Gray-scale}                                             &  \\ \cline{1-5}
\multicolumn{1}{|l|}{ISRTHDB}                                                                                      & \multicolumn{1}{l|}{ISRT, DU}                                                            & \multicolumn{1}{l|}{13,133}                                                        & \multicolumn{1}{l|}{277}                                                        & \multicolumn{1}{l|}{RGB}                                                    &  \\ \cline{1-5}
\multicolumn{1}{|l|}{\begin{tabular}[c]{@{}l@{}}BanglaLekha-\\ Isolated \cite{banglalekha} \end{tabular}}                              & \multicolumn{1}{l|}{-}                                                                   & \multicolumn{1}{l|}{20,319}                                                        & \multicolumn{1}{l|}{572}                                                        & \multicolumn{1}{l|}{Binary}                                                 &  \\ \cline{1-5}
\multicolumn{1}{|l|}{UIUDB}                                                                                        & \multicolumn{1}{l|}{UIU}                                                                 & \multicolumn{1}{l|}{576}                                                          & \multicolumn{1}{l|}{81}                                                         & \multicolumn{1}{l|}{RGB}                                                    &  \\ \cline{1-5}
                                                                                                                   &                                                                                          &                                                                                   &                                                                                 &                                                                             & 
\end{tabular}
\end{center}
\label{tab1}
\end{table}

\subsection{Train and Test Data}
The data set was split into train and test sets. The training set consists of 72,044 samples and the test consists of 13,552 samples. Apart from UIUDB data, all the dataset were split into 85\% training and 15\% test data. UIUBDB dataset in its entirety was considered for test set. The training and test set were named a, b, c, d, e and f. Table \ref{tab2} details the training and test datasets.

\subsection{Augmentation on Test set}

The test data was further augmented to simulate practical challenges of the task. Different augmentation techniques were applied to make the data varied and fluid. They are:

\begin{itemize}

\item Spatial Transformations: Rotation, Translation, Shear, Height/Width Shift, Channel Shift, Zoom.
\item Brightness, Contrast, Saturation, Hue shifts, Noise.

\item Occlusions.

\item Superimposition (to simulate the effect of text being visible from the other side of a page).

\end{itemize}

The augmented data was retrieved from the test set 'a' and 'c'. No augmentation was performed on the training set and to rest of the remaining test set. The number of samples for 'aug-a' and 'aug-c' are 2,168 and 2,106. Table \ref{tab2} details the augmented set with respect to the original non-augmented train and test split.

\begin{table}[tp]
\caption{Training and Test Data}
\begin{center}
\begin{tabular}{|l|l|l|l|l|}
\hline
Split name & \multicolumn{1}{c|}{\begin{tabular}[c]{@{}c@{}}Train-Test \\ Split\end{tabular}} & \begin{tabular}[c]{@{}l@{}}Training\\ Sample\end{tabular} & \begin{tabular}[c]{@{}l@{}}Test\\ Sample\end{tabular} & \begin{tabular}[c]{@{}l@{}}Test Sample\\ Augmented\end{tabular} \\ \hline
a          & 85\%-15\%                                                                        & 19,703                                                     & 3,490                                                  & -                                                               \\ \hline
b          & 85\%-15\%                                                                        & 359                                                       & 70                                                    & -                                                               \\ \hline
c          & 85\%-15\%                                                                        & 24,298                                                     & 4,381                                                  & -                                                               \\ \hline
d          & 85\%-15\%                                                                        & 10,908                                                     & 1,948                                                  & -                                                               \\ \hline
e          & 85\%-15\%                                                                        & 16,778                                                     & 2,970                                                  & -                                                               \\ \hline
f          & 0\%-100\%                                                                        & -                                                         & 495                                                   & -                                                               \\ \hline
aug-a      & 0-100\%                                                                          & -                                                         & -                                                     & 2,168                                                            \\ \hline
aug-c      & 0-100\%                                                                          & -                                                         & -                                                     & 2,106                                                            \\ \hline
\end{tabular}
\end{center}
\label{tab2}
\end{table}

\section{Bengali.AI CV Challenge 2018}
\subsection{Competition Evaluation Metrics}
Six different sources were used as test data such as test sets \textbf{a}, \textbf{b}, \textbf{c}, \textbf{d}, \textbf{e} and \textbf{f}. Additionally, two augmented data sets were produced from test set \textbf{a} and \textbf{c}. To account for the non-homogeneity of the sub-sets, we opted to use \textbf{Unweighted Average Accuracy} (UAA) as the evaluation metric for the competition. 
\begin{equation}
    Accuracy = \sum_{i=1}^{8} A_{i}
\label{eq1}
\end{equation}
In \eqref{eq1}, $A_i$ is the model accuracy on the i'th test sub-set.
We also compared the Weighted Average Accuracy (WAA) besides UAA in Table \ref{tab3}.

\subsection{Participating Teams and Corresponding Score}
The competition saw the largest amount of teams participating for any Computer Science related competition in Bangladesh. In total, 92 teams participated from 19 different institutions. Out of those, 57 teams submitted at least one or more viable models for Digit Recognition. The total number of entries for models were 685. Meaning, there were 12 model entries on average per team. The winning teams were \textbf{Backpropers}, \textbf{Digit\_Branch} and \textbf{Dekhi\_ki\_hoi}. Table \ref{tab3} lists the top 20 teams, their affiliations, corresponding score and number of entries. Note that the relative difference between the top two teams considering WAA is \emph{.01\%}.  As it can be seen that the 1st place winner had a staggering 54 algorithm submission. Digit\_Branch came 2nd submitting only one-third of that. Another fascinating thing is all 20 of the teams reached top-10\% Accuracy. Additionally, the top 3 teams achieved competitive scores with a difference of \emph{0.1\%} UAA.

\begin{table}[tp]
\centering
\caption{Standing of Top 20 teams}
\begin{center}
\resizebox{\linewidth}{!}{%
\begin{tabular}{|c|c|c|c|c|c|}
\hline
Ranking & \begin{tabular}[c]{@{}c@{}}Team\\ Name\end{tabular}& Affiliation  & \begin{tabular}[c]{@{}c@{}}Number\\ of Entries\end{tabular} & UAA & WAA \\ \hline
1& Backpropers& BUET  & 60 &0.99359 & 0.99484\\ \hline
2& Digit\_Branch& CUET  & 17 & 0.99296 & 0.99478\\ \hline
3& Dekhi\_ki\_hoi& BUET & 26 & 0.99177 & 0.99336\\ \hline
4& Sabbir Ahmed& BUET  & 55 & 0.9808& 0.98530\\ \hline
5& Sannin& DU  & 28 & 0.97889 & 0.98428\\ \hline
6& Lets Try& NSU  & 39 & 0.97606 & 0.98269\\ \hline
7& Diversense& KUET  & 28 & 0.96694 &0.97134\\ \hline
8& AUST\_Benzema& AUST  & 47 & 0.96188 & 0.97305\\ \hline
9& Kola& BUET & 19 & 0.95236 &0.96550\\ \hline
10& Rafizunaed& BUET & 8 & 0.94393 & 0.96777\\ \hline
11& Osprishyo& SUST & 17 & 0.93942 & 0.94547\\ \hline
12& Numta\_ai& BUET & 17 & 0.93631 & 0.94780\\ \hline
13& RUETvision& RUET  & 33 & 0.92723 & 0.94167\\ \hline
14& Eyes on you& CUET & 8 & 0.92701 & 0.93793\\ \hline
15& RUET\_13& RUET  & 12 & 0.92426 & 0.93662\\ \hline
16& Yellowchrom& DU & 11 & 0.91855 & 0.93691\\ \hline
17& Halum& KUET & 13 & 0.91619 & 0.93407\\ \hline
18& Sadman Sakib& BUET & 3 & 0.91387 &0.94281\\ \hline
19& Code\_crawlers& BUET & 10 & 0.91269 & 0.93004\\ \hline
20& Md Asadul Islam&UAlberta & 5 & 0.90296 & 0.91972\\ \hline
\end{tabular}%
}
\end{center}
\label{tab3}
\end{table}

\begin{figure}[b]
\includegraphics[width=\linewidth]{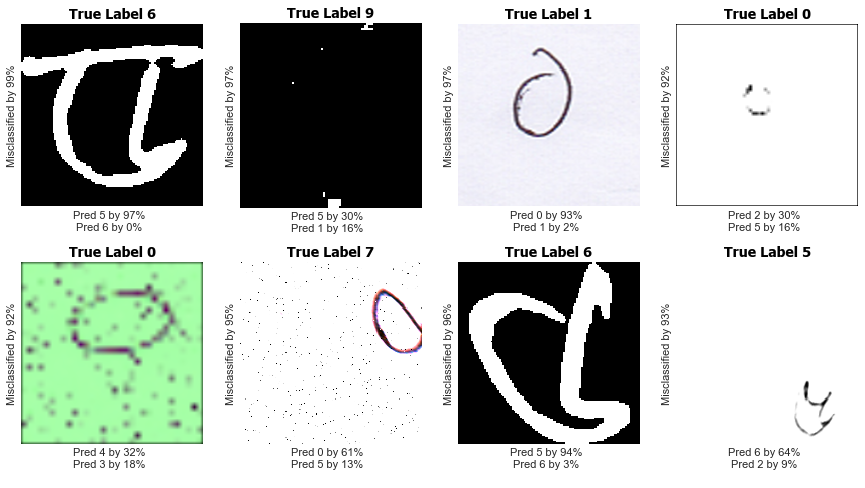}
\centering
\caption{Testing set images which were frequently misclassified by competition entries.}
\label{mislabeled}
\end{figure}

\subsection{\text{}Winning Algorithms}
\subsubsection{\textbf{Backpropers}}
The team used multiple architectures such as Resnet-34 and Resnet-50 \cite{resnet}. Transfer learning was performed using pre-trained weights and an ensemble architecture was made. Backpropers used PyTorch \cite{pytorch} deep learning library for creating the model and the whole training and validation schemes were executed in Google Colaboratory \cite{colab}. In total, they used 6 models to create an ensemble and averaged the score to get the final prediction. 

\subsubsection{\textbf{Digit\_Branch}}
Digit\_Branch used ResNeXt-50 \cite{resnext} as pre-trained weights for training their model. In total, 3 separate yet similar architectures were used. A data generation scheme called Overlay was used. It generated fixed 5,000 images from the training data. They excluded 0 and 4 from being added as a mirrored image in the Overlay. Those 5,000 images were used along with the original train set for training the three models. Data augmentation was performed such as Rotation, Shear, scale, Translation, Noise, Salt And Pepper, Contrast Normalization etc. 

\subsubsection{\textbf{Dekhi\_ki\_hoi}}
The entrant ensembled 4 different models and trained it end-to-end. The first model was Densenet-121 with pre-trained weights from Imagenet \cite{densenet}. The next two models had a combination of dilated convolution \cite{dilation} block stacked in different orders. Lastly, a repetitive plain convolution block was used to create the fourth model. The final prediction was averaged over from all four model's predictions. Data Augmentation was performed such as Gaussian Blur, Additive Gaussian Noise, Channel-wise random scaling, Affine scaling, Translation, Rotation and Shear, Contrast Normalization, and Additive Pepper noise.  

\begin{figure}[tp]
\includegraphics[width=\linewidth,trim={1cm 2.5cm 4.8cm 1.5cm}]{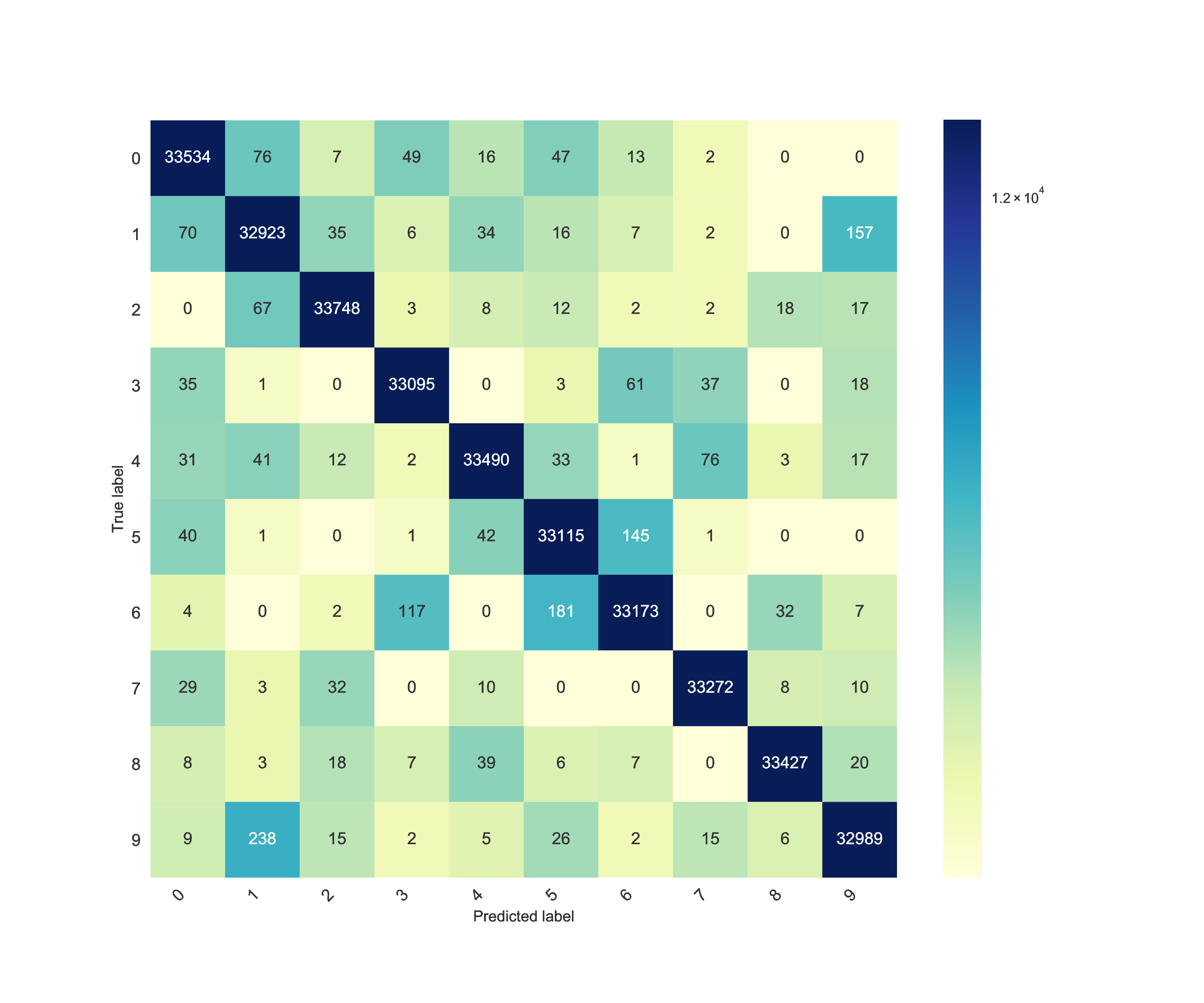}
\centering
\caption{Confusion matrix of competition entries surpassing 99\% UAA.}
\label{confmat}
\end{figure}

\begin{figure*}[!htbp]
\includegraphics[width=\textwidth]{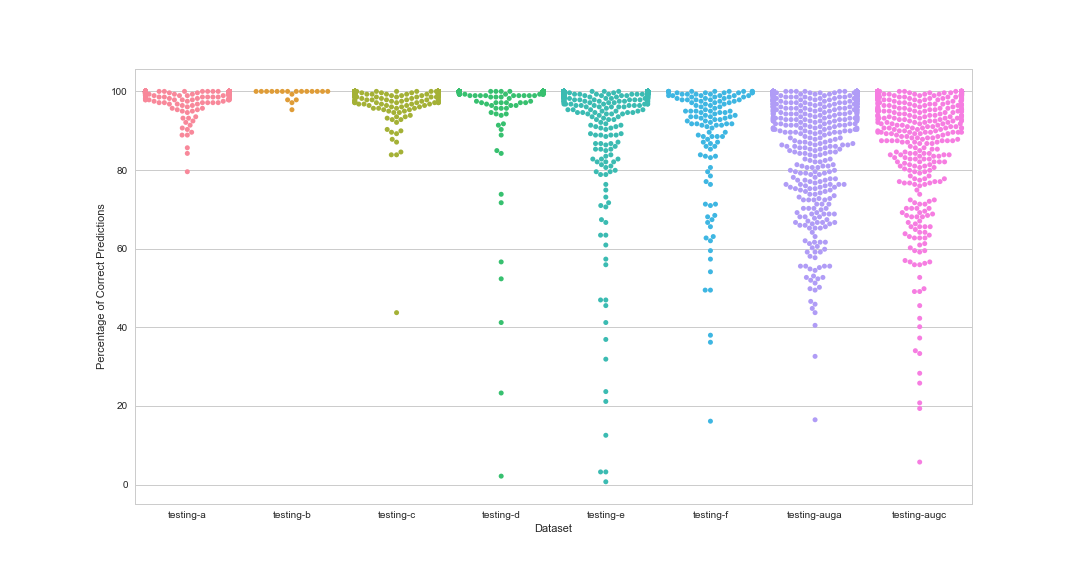}
\centering
\caption{Accuracy comparison of testing sets on top submissions.}
\label{comparison}
\end{figure*}

\section{Evaluating Label Noise in NumtaDB}
Large collections of labeled images have powered the recent advances in image classification. Large datasets are prone to having label noise both in the training and testing datasets \cite{labelnoise}. Deep learning models trained with large supervised datasets are robust to mislabeled data in the training set \cite{rolnick2017}. Noisy labels in the testing set could introduce biases which need resolution. To find out noisy labels, we manually inspected the frequently misclassified digits by the submissions in the Bengali.AI CV Challenge (Fig. \ref{mislabeled}). We found 69 mislabeled testing set instances among which 45 were from testing-e. The mislabeled digits have been replaced with correctly labeled data. The updated dataset is available at \href{http://bengali.ai/datasets}{http://bengali.ai/datasets}.

\section{Results Analysis}

To compare accuracy of models on different datasets, we took the top 335 submissions. We defined an accuracy metric of each sample as the percentage of models that predicted correct out of all submissions and calculated the distribution of accuracy of each sample point categorized by dataset.\\

As the data was assembled from various sources, the accuracy on testing sets varied with source. Analyzing the non augmented sets we found that testing-e was by far the most difficult followed by testing-f. This was expected as testing-e and testing-f were collected manually and so a lot of samples naturally contained noise and deformations. In contrast, the other testing sets were much easier to decipher for the models with testing-b being the easiest. Testing-auga and testing-augc has similar error distributions and accounted for the majority of the error (Fig. \ref{comparison}).\\
An analysis of the influence of different manner of augmentations was also done. This information is invaluable during training as it can point out where the most focus is needed when augmenting the training set. We used a combination of several augmentations (translate, coarse dropout, perspective transform, hue and white balance shift, pixelation, superposition, contrast normalization, salt pepper noise addition, Gaussian blur, rotate and shear). Our augmented test datasets had a total of 385 unique combinations of different kinds of augmentations. In our analysis of the augmented testing sets testing-auga and testing-augc, we calculated the average number of incorrect predictions made by the top 279 submissions all of which scored above 90\% accuracy. We found that the combination of perspective transform, gaussian blur and shear was the hardest for models to predict with an average of 213 incorrect predictions per sample. Combinations of shear, hue - white balance shift and perspective transform was the next hardest with 205 mistakes per sample. All other combinations had an average incorrect prediction per sample less than 149.

\section{Discussions \& Conclusion}
Bengali.AI Computer Vision Challenge saw a large number of teams participating with hundreds of algorithms to clinch the title of the best Bengali handwritten digit recognition model. Adamant enough, they strove to explore unknown horizons by implementing both state-of-the-art and traditional machine learning techniques to come out victorious. Through this competition they gathered abundant information regarding competitiveness, practical challenges as well as solving computer vision related problems. This challenge is the first of a series of competition that will try to illuminate both the beginner and veteran artificial intelligence practitioners to solve real life issues using machine learning and beyond.

\section*{Acknowledgment}
Bengali.AI is a community working to solve the scarcity of standardized datasets for Bengali Natural Language Processing and Computer Vision research. We are grateful to all the members of the community who took part in data contribution and the successful organizing of the event. Bengali.AI Computer Vision Challenge was exclusively sponsored by \href{http://www.apurbatech.com/}{Apurba Technologies Inc.}
\balance
\bibliographystyle{unsrt}
\bibliography{reference}

\end{document}